# Cross-language Framework for Word Recognition and Spotting of Indic Scripts

[a]Ayan Kumar Bhunia, [b]Partha Pratim Roy*, [a]Akash Mohta, [c]Umapada Pal

[a]Dept. of ECE, Institute of Engineering & Management, Kolkata, India
[b]Dept. of CSE, Indian Institute of Technology Roorkee India
[c]CVPR Unit, Indian Statistical Institute, Kolkata, India
[b]email: proy.fcs@iitr.ac.in, TEL: +91-1332-284816

## Abstract

Handwritten word recognition and spotting of low-resource scripts are difficult as sufficient training data is not available and it is often expensive for collecting data of such scripts. This paper presents a novel cross language platform for handwritten word recognition and spotting for such low-resource scripts where training is performed with a sufficiently large dataset of an available script (considered as source script) and testing is done on other scripts (considered as target script). Training with one source script and testing with another script to have a reasonable result is not easy in handwriting domain due to the complex nature of handwriting variability among scripts. Also it is difficult in mapping between source and target characters when they appear in cursive word images. The proposed Indic cross language framework exploits a large resource of dataset for training and uses it for recognizing and spotting text of other target scripts where sufficient amount of training data is not available. Since, Indic scripts are mostly written in 3 zones, namely, upper, middle and lower, we employ zone-wise character (or component) mapping for efficient learning purpose. The performance of our cross-language framework depends on the extent of similarity between the source and target scripts. Hence, we devise an entropy based script similarity score using source to target character mapping that will provide a feasibility of cross language transcription. We have tested our approach in three Indic scripts, namely, Bangla, Devanagari and Gurumukhi, and the corresponding results are reported.

# 1. Introduction

Handwritten word recognition has long been an active research area because of its complexity and challenges due to a variety of handwritten styles. There exist many research works towards handwritten word recognition in Roman [1, 6, 17], Japanese/Chinese [2, 3] and Arabic scripts [5]. To overcome the drawbacks of recognition approaches, word spotting technique [11, 14, 23, 25] is used for information retrieval purpose. Word spotting was developed as an alternative approach for knowledge retrieval from document images which avoids conventional recognition framework. Researchers have created numerous public datasets in different scripts for developing tasks, such as, word recognition, word retrieval, etc. [21, 24, 25].

Although a number of investigations have been made towards the recognition of isolated handwritten characters and digits of Indian scripts [8, 39], only a few pieces of work [7, 8, 17, 20] exist towards offline handwritten word recognition in Indian scripts. Recognition of Indian scripts [22, 38] is difficult due to their complex syntax and spatial variation of the characters when combined with other characters to form a word. Modifiers are formed when vowels are connected to the consonant and these modifiers are placed at the left, right (or both), top or bottom of the consonant. Presence of 'Matra' and 'modifiers' [22] makes the recognition and spotting tasks of Indian script more difficult as compared to other non-Indian scripts. Hence most of the existing word recognition works in Indic script are performed based on the segmentation of characters from words.

Dataset is a necessary and important resource which is required to develop any recognition system for benchmarking. It has been observed that the availability of training data for handwritten task of each Indian script is not uniformly distributed i.e. some scripts like Bangla, Devanagari, Tamil etc. have a lot of data compared to other scripts/languages in India. Generation of synthetic data was also attempted to increase the size of dataset [37]. Most of the text recognition systems available in Indic scripts are performed in few popular scripts like Bangla, Devanagari, Tamil, etc. Due to the lack of proper datasets, research in other Indic scripts is not progressing.

To overcome the unavailability of datasets, researchers from the speech recognition community have developed speech recognition systems for low-resource languages with the help of large datasets of known language using cross language framework [9, 10]. There exist many pieces of work on cross language speech recognition. Generally, in these works, phoneme to phoneme mapping technique has been applied. Till date, several experiments have been performed in the field of cross language speech recognition. The spectral feature (like Mel Frequency Cepstral Coefficient (*MFCC*)) based baseline system for Mandarin and Arabic automatic speech recognition (ASR) performance was outperformed by using feature extraction from an English-trained MLP in [12]. In [15] English-trained phone and articulatory feature MLPs was considered for a Hungarian ASR system to study the cross-lingual portability of MLP features from English to Hungarian. The work presented in [9], describes the development of a Polish speech recognition system using Spanish as the source language. This was done by using cross-language bootstrapping and confidence based unsupervised acoustic model training, in a combined manner. Also, in [10], it was proposed that to address lack of training resources, data from multiple languages can be used. Here a multi-language Acoustic Model (AM) was directly applied, as a language independent AM (LIAM) to an unseen language, considering limited training resource of target language.

Inspired with the success in speech recognition, we attempt cross-language handwritten text recognition in this paper. Cross language handwriting recognition refers to the process of word recognition of a target language, by a system which has been already trained by different (source) languages. In this paper, we propose a novel approach of cross language handwritten word recognition by source to target language character mapping technique. To our knowledge, the task of cross language handwriting recognition has not been performed earlier. To address this problem, we propose a method in which the character models, trained using source language, are used for recognition of target script. The training models created from source language are used for mid-level transcription of target language. Next, a character mapping from source language to target language is performed to obtain the final transcription. Similarly the source language character models are used for word spotting in target language.

Our proposed cross-language framework uses the zone segmentation concept [20] in order to reduce the number of unique characters in Indic scripts. The major contributions of this paper are the following: 1) Use of cross language framework for word recognition and spotting: Although there are quite a few pieces of works in cross language speech recognition, the idea of cross language handwritten word recognition or spotting is a novel attempt in the handwriting recognition community. 2) Target to source script mapping using majority voting: We propose a character mapping method in order to find a mapping between source to target characters. 3) Script similarity score calculation: A novel script similarity measure is proposed to evaluate the similarity between source and target scripts. This central idea of cross-language framework is general and can be extended to other low resource scripts where enough training date is not available. The proposed paradigm will help in developing recognition and spotting approaches for low resource scripts. There exists not much handwritten recognition work on Indic scripts such as Gurumukhi, Oriya, and Assamese etc. Thus, developing an efficient cross language will be useful for such low resource scripts using large resource scripts.

The rest of the paper is organized as follows. In Section 2, we describe the similarities among Indic scripts. The advantaged of zone-wise word division for cross-language similarity is explained. We have reviewed the zone segmentation method [20] in Section 3. In Section 4, we detail our proposed framework on word recognition and spotting using cross-language scripts learning. In Section 5, the script similarity score computation approach is explained. We demonstrate the performance of our framework in Section 6 with different Indic scripts. Finally, conclusions and future work are presented in Section 7.

## 2. Similarity in Indic Scripts

The root of most Indian scripts is Brahmi. Over the years, Brahmi has slowly transformed into popular modern scripts namely Bangla, Devanagari, Gurumukhi, Gujarati, etc. This may be due to their same origin and successive evolution of the characters used in different parts of the country has resulted in the origin of new scripts. Most Indian languages are also descended from ancient Sanskrit language. Because of a single origin, the character names in many scripts, like Devanagari, Bangla, and Gurumukhi etc., are similar and shapes of the characters share similar appearance [22]. This can be elaborated by considering an example of the same character from three different scripts like the character '্ম'(Bengali), ਸ(Gurumukhi) and म (Devanagari). Though they belong to different scripts, they are similar in appearance. The successive evolution of different major Brahmic scripts in India and Southeast Asia is discussed in [27]. It is mentioned that Bangla, Devanagari, Gujarati and Punjabi scripts have some shape similarities. Among the south Indian scripts, Kannada has some similarity with Telugu, and Malayalam has similarity with Tamil. Due to these similarities, proposals have been made to develop a general OCR engine encompassing all the scripts. On the other hand, the similarities create more difficulty in script identification tasks. Among these similar scripts, one or two scripts are used in communication by a large section of the country. Hence, OCR systems of such dominant scripts are being developed in recent years. But, many scripts are still remained unexplored due to lack of proper dataset. For transcription of such unexplored scripts, labelled training data is hard to get.

Most of the Indic scripts are written from left to right. Unlike Latin, character-modifiers of Bangla, Devanagari, Gurumukhi and some other scripts are attached to the consonant (that appears only in middle zone) in any of the 3 zones: upper, middle or lower zones. Fig.1. shows an example of a Bangla word image and its 3 different zones. In these scripts, characters usually have "*Matra*" to which they are attached at the top. Often in such scripts, if a consonant is followed by a vowel, a vowel symbol is added to the consonant either to the left, right, top or bottom, depending upon the usage. A consonant or a vowel following a consonant sometimes takes a compound orthographic shape, which we call as compound character.

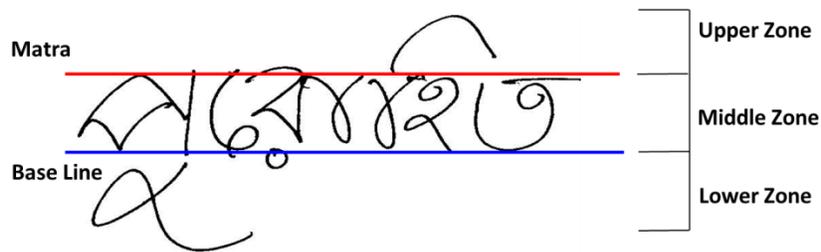

**Fig.1.Example of a Bangla word showing the 3 different zones**

F
o
r

i
n
s
t
a

characters sit side by side to form a word, corresponding *Matra*s generally touch and generate a long line.

**(a)**

**(b)**

**Fig.2. Some examples show similarity among (a) characters and (b) words in Bangla, Devanagari and Gurumukhi scripts. Similar portions of the characters in three scripts are marked in red dotted line.**

In spite of having much similarity among the characters of different Indic scripts, the recognition task becomes difficult due to appearance of characters in three different zones in Indic script during word formation. It is observed that the presence of modifiers reduces the cross language similarity more than the simple consonant characters which are situated only in middle zone. Although some of the consonant characters of the three scripts are structurally similar in appearance as discussed in the previous sections but when modifiers are attached to these characters, the shapes of the resultant characters differ to a greater extent in different scripts. A diagrammatic illustration of this fact is shown in Fig. 3. The similarity among modifiers is much less than the middle zoned characters among the scripts. Hence, we have used zone-wise word components of source scripts to train and the trained models are used to test zone-wise word images of target script. When the source script character models are applied to target word image, we obtain the transcription of target word using source script characters. We refer it as mid-level transcription. This mid-level transcription is then converted to the target script using character mapping.

|  | Bangla | Gurumukhi | Devanagari |
|---|---|---|---|
| Full Zone | কু | ਬੁ | कृ |
| Middle Zone | ক | ਬ | क |
| Lower Zone | ু | ੁ | ृ |

|  | Bangla | Gurumukhi | Devanagari |
|---|---|---|---|
| Full Zone | কী | ਬੀ | की |
| Middle Zone | কা | ਬਾ | का |
| Upper Zone | ী | ੀ | ी |

<div align="center">(a)            (b)</div>

**Fig.3. Example showing that the maximum similarity lies in the middle zoned character among the scripts. Similarity of the characters in a single (same) zone is much significant than combining other zones, e.g., (a) with lower zone (b) with upper zone.**

## 3. Zone Segmentation in Indic Scripts

Zone segmentation plays an important role in Indic script word recognition as proposed in [20]. As mentioned earlier, characters of most of the Indic scripts are written in upper, middle and lower zones. With morphological combination of characters with modifiers, the number of character classes becomes huge [20]. To overcome the problem of large number of characters classes, zone-wise character segmentation and combination approach significantly reduces the combination of characters with the modifiers. It was shown that zone-wise recognition method significantly improves the word recognition performance than conventional full word recognition system in Indic scripts [13, 20]. To make this document self-contained, we have briefly reviewed the zone segmentation method that was introduced in the works [13, 20].

To perform the zone segmentation approach in Indic script, the first step is to detect the proper region of Matra which is a challenging task due to complex writing style. Unlike printed word [22] where the row with the highest peak in horizontal projection analysis detects the Matra, it is rarely true in cursive handwritten words. The zone segmentation approach due to Roy et al. [20] showed good performance in Indic scripts. We have used similar approach for segmentation of the three zones in Indic scripts. A rule based approach was considered to detect the approximate location of Matra line. For this purpose, three different information namely, highest peak of horizontal projection, regression line of depth-points of water reservoirs and the projection profile in the upper half of the word are considered. Next, a window of *Matra* region is considered to find the upper zone components. The skeleton-segments of the region are analysed to check if some of the segments are moving in the upward direction from the Matra region. The segments which move upwards are considered as upper-zone components of the word image.

Lower zone components are also segmented from the word image. The lower zone segmentation by projection analysis does not perform well because of the irregular size of the characters within a word. To perform better segmentation of lower zone components a shape matching based algorithm is introduced [20]. Modifiers are searched in lower portion of the word image using a shape matching algorithm. The touching locations of lower zone modifiers are found out by the analysis of the skeleton of the image. If the residue shape components below those touching locations are matched with any of the lower zone characters with high matching confidence, those portions are separated from middle zone. Fig.4 shows examples of zone segmentation on Bangla, Devanagari and Gurumukhi.

| Script | Full word | Middle zone | Modifiers |
|--------|-----------|-------------|-----------|
| Bangla |  |  |  |
| Devanagari |  |  |  |
| Gurumukhi |  |  |  |

**Fig.4. Examples showing zone segmentation on Bangla, Devanagari and Gurumukhi word images.**

## 4. Proposed Cross-Language Framework

We have used our proposed cross language technique in two state-of-the-art frameworks: word recognition and spotting from handwritten word images. Our technique is based on the character mapping between source and target scripts. The mapping of characters from source to target scripts depends on similarity. Fig.5 describes the architecture of the overall framework of the proposed system. The more the scripts are similar the recognition and spotting performance will yield better results. In this Section we discuss the cross-language recognition and spotting framework in details. The similarity index between two scripts is detailed in next Section.

For word recognition and spotting tasks, we have trained the character models using available data from source script. Next, these character models are used for target script word recognition and spotting tasks. The obtained recognition result is mid-level transcription, i.e. in the form of character sequence of source script. The mid-level transcription is then converted to target script using source to target character mapping. For word spotting, firstly, the query keyword of the target script (testing) is mapped to character sequence of source script. Then this mid-level query keyword is used to search similar words in target text lines. Due to the similar morphology of Bangla, Devanagari and Gurumukhi character set we have adopted the zone-wise matching approach in our framework.

In the subsequent subsections, we describe our proposed methodologies in details. In Section 4.1, we describe the feature extraction process for character modeling of source and target scripts. In Section 4.2, the mapping from source to target character is explained. This mapping is used for both cross language word spotting and recognition. Finally, we detail the complete framework for cross language word recognition and word spotting in Section 4.3 and Section 4.4, respectively.

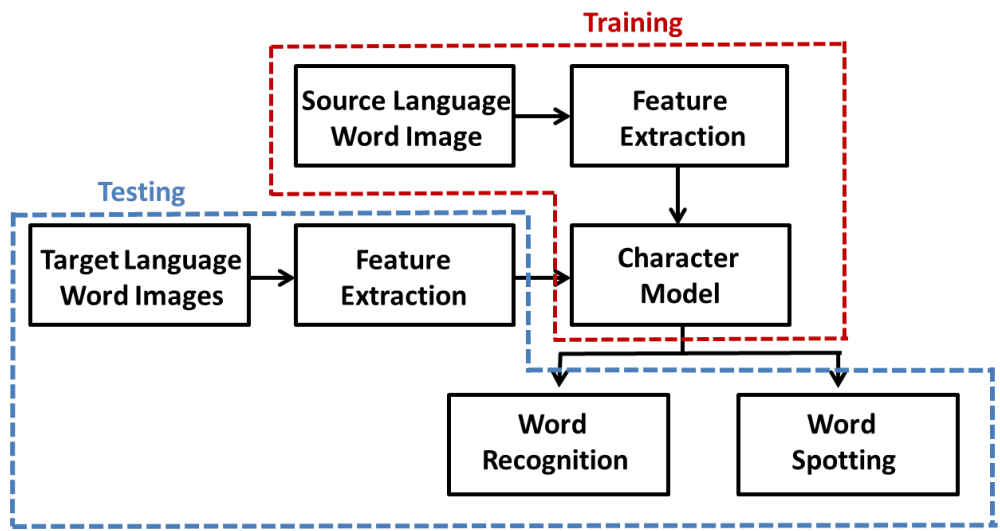

**Fig.5. Proposed architecture of the cross-language technique**

## 4.1. Character Modeling using Source Scripts

Training of source script characters are performed using zone-wise components. Since, components in middle zone are cursive and touching they are trained using Hidden Markov Model to avoid segmentation of touching characters. For isolated components in upper and lower zones, SVM is used for corresponding component modeling. In both HMM and SVM classification, Pyramid of Histogram of Gradient (PHOG) feature has been used [13, 20]. In PHOG feature extraction approach, an image is divided into cells at several pyramid level and from each level (i.e. $N$=0, 1, 2,..), histogram of oriented features are extracted. In this work, with 2 levels of resolution, we obtained (1×8) + (4×8) + (16×8) = (8+32+128) = 168 dimensional feature vector for individual sliding window position.

**A. Middle-zone component modelling using HMM:** The middle-zone word components from source script are considered for HMM [21] training. Except cursive and touching behavior of handwriting, another major reason behind choosing HMM is that it can model sequential dependencies. From middle-zone word image, a sliding window is moved from left to right direction with an overlapping. PHOG feature is extracted from each position of the sliding window. Next, training is performed using continuous density HMM [26].

**B. Upper/Lower zone modifier modeling using SVM:** The isolated components which are included in upper and lower zones are segmented using connected component (CC) analysis [36] and next they are recognized and labelled as text characters. After resizing the images to 150x150, here also PHOG feature of vector length 168 is extracted from the components of upper and lower zone modifiers. Next, Support Vector Machine (SVM) classifier [18, 32, 33] has been used to classify these components. Radial Basis Function (*RBF*) kernel is used in our experiment study to classify upper and lower zone modifiers (as shown in Fig.6).

| Scripts | Upper zone modifier | Lower zone modifier |
|---|---|---|
| Bangla | 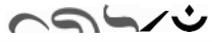 | 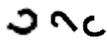 |
| Devanagari | 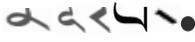 | 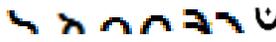 |
| Gurumukhi | 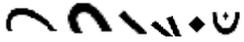 | 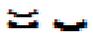 |

**Fig.6. Examples showing upper and lower zone modifiers in Bangla, Devanagari and Gurumukhi scripts.**

## 4.2. Look-up-Table from Source to Target Characters

It is noted that similarity among characters in corresponding zone is much higher than considering the full zone information. The evaluation of script similarity is detailed in Section 4. In experiment section (Section 5), advantage of zone-wise character similarity is discussed.

To utilize this script similarity, we have used character mapping procedure using majority voting. For this purpose, each zone-wise character component of target script is first recognized using the source script character models. During this recognition step, samples of a target character component may get recognized by more than one source script characters. It is due to non-availability of similar-shape character in source script. The source character by which the recognition is performed maximum is chosen for mapping, i.e., we select the source script character which appears highest number of times as recognition label of that target script character. The scheme of majority voting is explained in Fig. 7. The source character whose frequency is more is chosen for recognition. Character mapping helps us in replacing the character of the target script with that of the source script. For this purpose, few samples of each character of target script are required. An example of lookup table for Bangla as target script and Devanagari as source script is given in Table I.

**Table I: Look-up table used for Bangla as target (testing) and Devanagari as source (training) scripts.**

| Target Script (Bangla) | Source Script (Devanagari) | Target Script (Bangla) | Source Script (Devanagari) | Target Script (Bangla) | Source Script (Devanagari) |
|---|---|---|---|---|---|
| ক | क | থ | थ | য | य |
| গ | ग | ন | न | স | स |
| ঘ | घ | ব | व | ল | ल |
| ড | उ | ম | म | হ | र |
| থ | ध | প | प | অ | अ |
| জ | ज | ফ | फ | ঙ | ङ |
| ঠ | ठ | ষ | ष | দ | द |
| ঝ | बा | ঢ | ढ | ণ | ण |

**A. Character mapping for middle zone:** Each middle zone component of target script is recognized as one of the source script character. The recognition is performed using HMM and labeled by one of the source characters. Finally, a set of similar target characters is recognized and a majority voting based decision is considered for that target character class. In Fig.7 (a) Bangla character 'ক' and Gurumukhi character 'ਦ' are being recognized as different individual Devanagari (source script) character and next final mapping is performed based on majority voting.

**B. Character mapping for modifiers:** As mentioned earlier, zone-wise character similarity among Indic scripts is more prominent, hence we use upper and lower zone modifiers separately for character mapping. Since, SVM was used for classification purpose; we have used this classifier for modifier mapping between source and target scripts. Upper and lower zone modifiers are mapped using SVM model trained by source script modifiers. Some examples of modifier mapping in both upper and lower zones are shown in Fig.7 (b).

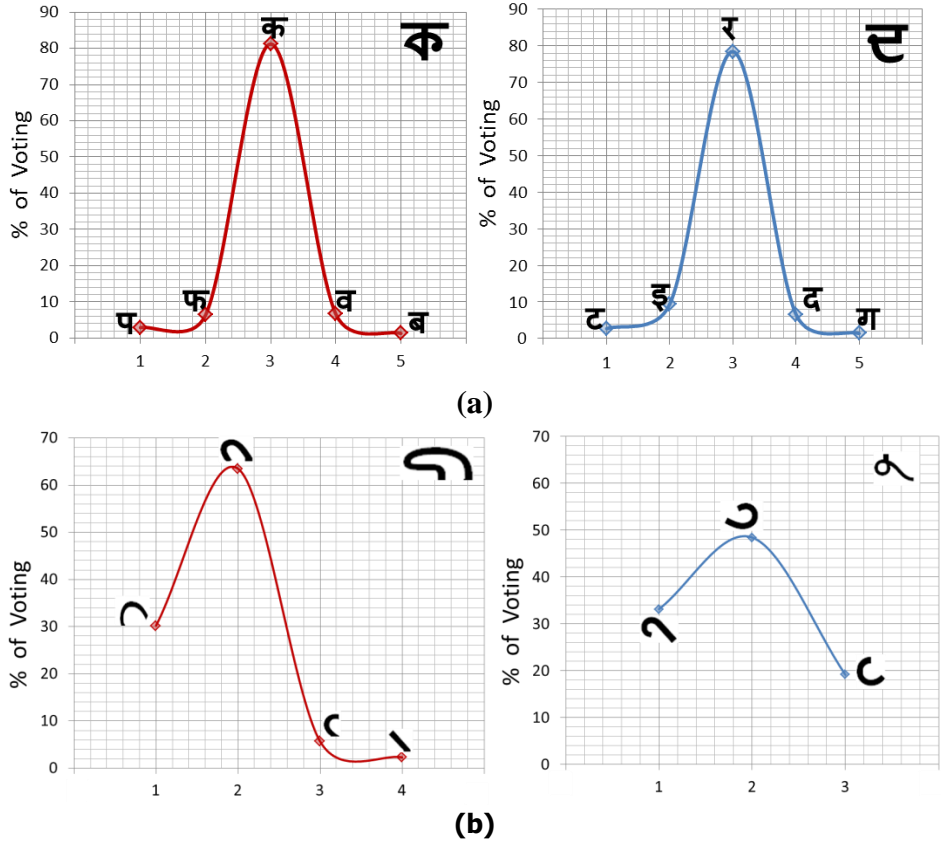

**Fig.7.Majority voting examples using (a) Bangla character 'ক' and Gurumukhi character 'ਦ' '(b)Bangla modifier 'ো' and Bangla modifier 'ৌ'. The y-axis shows the frequency of the recognition accuracy from source character to target character.**

### 4.3. Cross-Language Word Recognition

After creating the character look-up-table from source to target characters, the word images from target scripts are recognized using cross-language framework. Here, given a word image from target script, zone segmentation is performed to separate middle zone portion and modifiers. Zone segmented word image is recognized using zone-wise classifier models trained from corresponding source script. Middle zone components are recognized using HMM, and modifiers in upper and lower zone are recognized using SVM. The recognition results thus obtained would be character sequences from source script. Hence, these labels of zone-wise components are mapped to target script characters using Source-to-Target Look-Up-Table (LUT). Then full word recognition is being done combining the results of middle zone and modifiers. The detailed block-diagram of the propose cross-language recognition system is given in Fig.8.

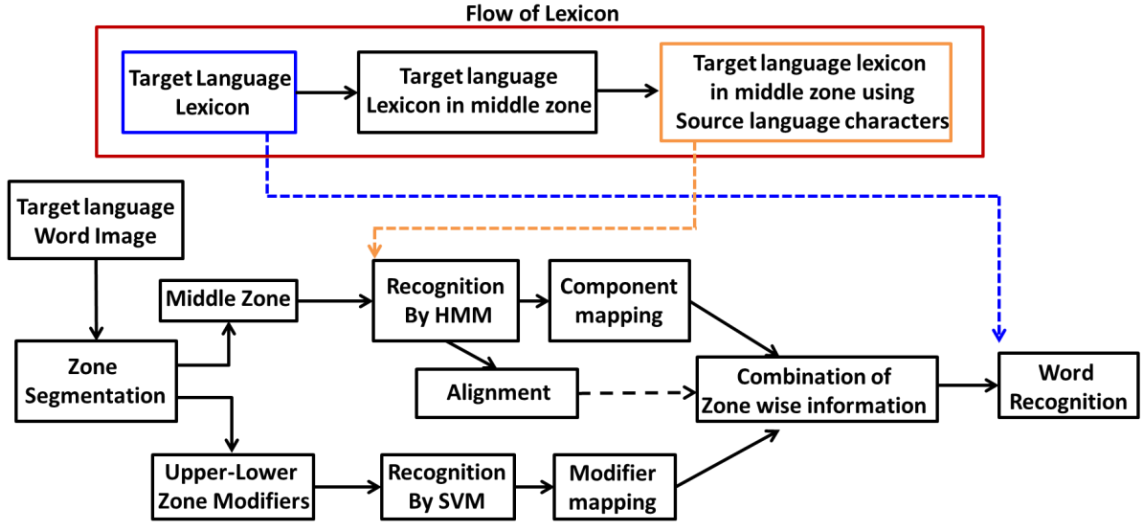

**Fig.8. Detailed flow diagram of cross-language word recognition.**

### 4.3.1. Lexicon preparation for target script word recognition

To recognize the target scripts of a set of lexicon ($L^T$) using zone-wise combination, the lexicon needs to be modified. Since, the HMM of our framework is lexicon-based the middle zone components need to be recognized with a lexicon consisting middle zone components only. The lexicon modification for middle zone components is performed in two-steps.

1. Lexicon ($L_M^T$) containing middle zone components only from $L^T$.

2. A mapped lexicon ($L_M^S$)$_{LUT}$ of $L_M^T$ from source script by using LUT.

In the first step, the lexicon ($L^T$) of the target script is converted to its equivalent transcription containing middle zone components only. Here, the upper and lower zone characters are avoided to make lexicon ($L_M^T$) containing middle zone characters only. In second step, middle-zone characters of target lexicon are mapped to the transcriptions of source script by character replacement. The character mapping is performed using Look-up-Table as discussed in Section 3.3. We call this lexicon ($L_M^S$)$_{LUT}$ as Mid-Level target to source lexicon or simply mid-level lexicon.

The actual target lexicon ($L^T$) is used in final word recognition result and the mid-level lexicon is used in middle zone recognition during HMM. Fig.9. shows the translation of lexicon from ($L^T$) to $(L_M^S)_{LUT}$ through intermediate step of ($L_M^T$).

| Meaning | Utility | Example - 1 | Example - 2 | Example - 3 |
|---|---|---|---|---|
| Target Language Lexicon [ ($L^T$) ] | Used in full word recognition using modifier combination | দিনগুলি | কবিগুরু | কমলিনী |
| Target language Lexicon in middle zone [ ($L_M^T$) ] | - | দিনগাল | কাবগব | কমালনা |
| Target language lexicon in middle zone using Source language characters [ $(L_M^S)_{LUT}$ ] | Used in middle zone recognition using HMM | দিनगाल | कावगव | कमालना |

**Fig.9. Example showing translation of lexicon along with its utility.**

**4.3.2. Lexicon based middle zone word recognition and alignment:** During middle-zone word mapping, one-to-one mapping is a trivial one, but problem arises when two or more target script characters are mapped to same source-script character. To solve this we adopted a lexicon-based middle-zone matching method for target script. For, one-to-many mapping situation, a single label of source character is replaced by a probable target script character serially and generated word is searched in the lexicon. If there exists any result in the lexicon, then that word is chosen as middle-zone recognition result. An illustration is given graphically in Fig.10.

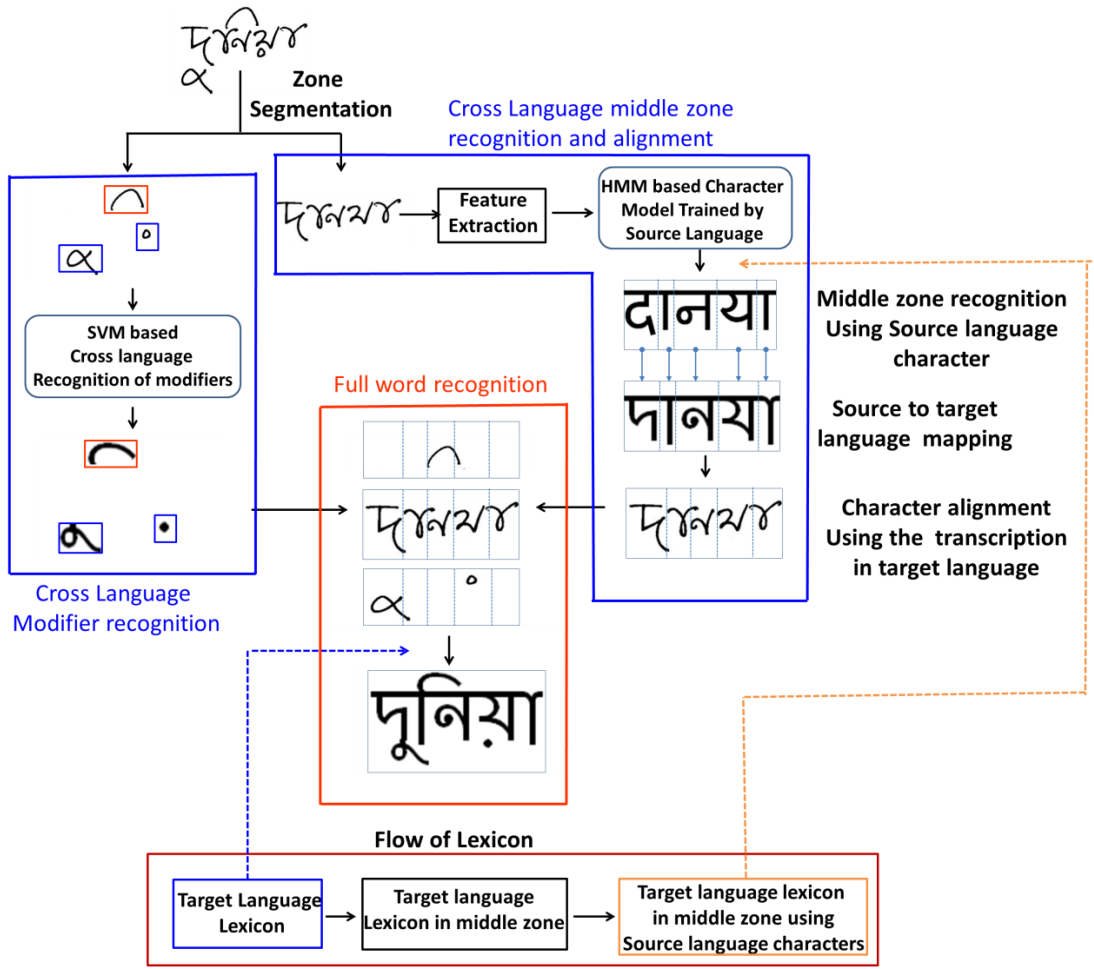

**Fig.10. Example showing alignment for zone segmented word 'दुनिया' where Devanagari is considered as training/source and Bangla as testing/target.**

### 4.3.3 Full word recognition by combining zone-wise information

After computing the zone-wise recognition results (upper and lower zone modifiers are recognized by SVM and middle zone characters by HMM of a word (X) and recognized character labels are obtained) the labels of upper and lower zones are associated with labels of middle zone. Details of the upper lower zone information combination along with the middle zone result are provided in [20]. The association of character labels can be considered as a path-search problem to find the best matching word where each character label will be used only once. For estimating the boundaries of the characters in the middle zone of a word, Viterbi Forced Alignment has been used in the middle-zone of the word. With the embedded training of FA, the optimal boundaries of the characters of the middle-zone are found. After obtaining the character boundaries in the middle zone, the respective boundaries are extended in the upper and lower zones to associate characters present in upper and lower zones with the middle zones characters. Similarly, we generate N such hypothesis using N-best Viterbi list obtained from middle zone of the word. Now among these N-best choices, the best hypothesis is chosen combining upper and lower zone information discussed as follow.

A middle zone character generally is associated to its corresponding upper and lower zone modifies. But, due to complex handwriting styles, some upper/lower zone modifiers may not appear exactly above and below of their middle zone character. To handle such situations, the association rule is made flexible. A middle zone character is associated not only to its exact upper and lower zone modifiers but it can also associate with one modifier with previous or next modifier from upper and lower zone. For each word a set of associated words may be obtained. Each associated word is matched with the lexicon (*L*) and the best matched associated word is the combined zone-wise result of the word. The similarity score in lexicon matching is obtained using string edit distance. Thus, we obtain a distance score for each associated word along with its word selected from lexicon. The scores are next sorted and the lexicon word with minimum score is considered as best result. We refer the work [20] for more details about the association rule.

### 4.4. Cross-Language Word Spotting

Cross language word recognition framework uses a lexicon matching based approach which creates a problem when our number of query words is a bit high. Here we adapt our cross language framework for lexicon free information retrieval through word spotting by HMM based scoring [19]. The word spotting procedure is shown in Fig.11.

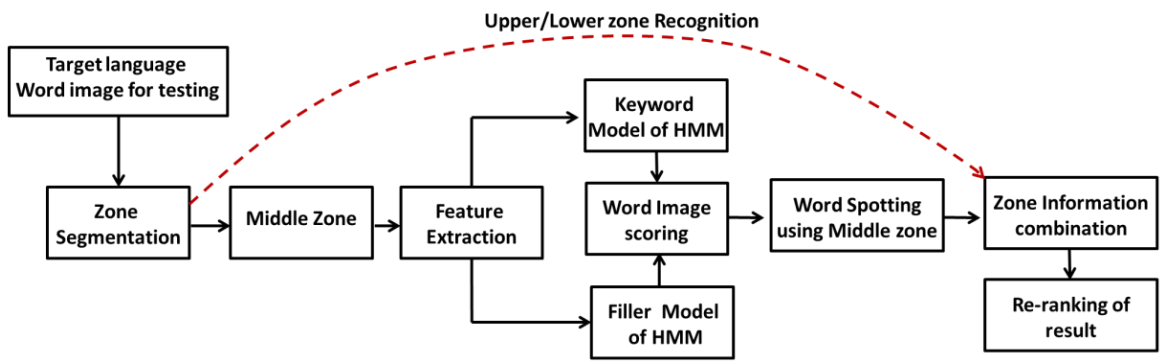

**Fig. 11. Word spotting procedure using zone segmentation.**

**Word image Scoring using HMM:** Firstly, zone segmented word images from source script are used to train the character model of HMM. The probability of the character model of the text line is then maximized by Baum-Welch algorithm assuming an initial output and transitional probabilities. Using the character HMM models, a filler model [19] has been created which is shown in Fig.11. The filler model represents isolated character models. During training phase, character HMMs are trained for each character of the alphabet based on transcribed word images from source script. At the recognition stage, the trained characters HMMs are connected to a keyword text model in order to calculate the likelihood score of the input word image. This likelihood score is finally normalized with respect to a general filler model before it is compared to a threshold. The score s(X) of image X for keyword W is based on posterior probability P(W|X). From Bayes' rule it follows that

$$log\ p(W|X)\ =\ log\ p(X|W)\ +\ log\ p(W)\ -\ log\ p(X) \qquad (1)$$

The prior $p(W)$ can be integrated into a keyword specific threshold that is optimized at training stage. For arbitrary keywords that are not known at the training stage, we assume equal priors. $p(X|W)$ is modelled using a HMM and $p(X)$ is modelled using a Filler model. The score S(X) is then compared with a threshold $T$ for word spotting.

$$S(X)\ =\ log\ p(X|W)\ -\ log\ p(X) \qquad (2)$$

The optimal value of $T$ can be determined in the training phase with respect to the user needs.

**Word Spotting using zone wise information combination:** Middle zone segmentation is very much effective to train the HMM character model because of significant reduction in the number of character sets. Zone-wise segmentation approach increases the word spotting performance for Indic scripts significantly. As we have adopted the approach of zone segmentation for our cross language word spotting framework, it demands a two-step mapping technique while searching for the testing target word images. In the first step, the ASCII keywords given by user (query keyword) are mapped to middle-zone based keywords. To do so, each character from full word level is mapped to middle zone level by a function according to a set of rules, e.g., the character modifiers "কৌ" and "কু" will be "কৌ" and "ক" respectively. In the next step, the middle zoned characters of the target script are mapped to source script character which will be used to generate the keyword model of HMM. Here, the same mapping rule used in word recognition (discussed in Section 3.3) is applied. Few examples are shown in Fig. 12 for this two-step mapping.

| Query Keyword of Target language | Middle zone mapping | Target to source Mapping (query keyword For HMM) |
|---|---|---|
| সাইনবোর্ড | সাহনবোড | सारनावाड |
| অনুরাগ | অনবাগ | अनवाग |
| কলিযুগ | কালযগ | कालयग |

**Fig.12: Example showing the two step mapping procedure from target script to source script through zone segmentation. (In this example we considered Bangla as target and Devanagari as source.)**

Due to reduced content from middle zone, the text component from middle zone may match with other words. For example, different words like 'চলি', 'চালু', 'চাল', 'চৌল' will be reduced to 'চাল' using middle zone portion. Because, the middle zone portion of this word image is segmented, the distinguishing features from upper and lower zones are neglected. So, searching with similar middle-zone mapped keywords will provide false positives which need special care while searching. Other zone-wise information will be useful to overcome this problem. Hence, combination of word-spotting performance using zone-based information is used in our system to overcome the shortcoming of middle-zone based spotting system. By combining zone information, the zone-wise information will complement each other for word retrieval. The combined system will be used for re-ranking the retrieval result obtained from middle zone based approaches. To combine the zone-wise information, we first recognize the upper lower zone modifier using the SVM-model trained by source script modifiers. Then Viterbi algorithm is used to get the character boundaries with in the testing word image. The recognition result of the upper-lower zone modifiers are combined along with its positional information for re-verification. For every searching keyword word, we have a look-up table containing the information about the modifiers. After recognizing the upper-lower zone modifiers of the testing word image, results are compared with that look-up table. Thus we eliminate the false positive cases obtained after using only middle zone information. In this re-verification context, we want to mention that even if we avoid the exact recognition results of the modifiers, we may use the information regarding number of modifiers in each zone of the testing word image and compare it to query keyword correspondingly. That also provides another way to upper-lower zone information combination approach to re-verify the results after middle zone based word spotting. Fig. 13 gives the diagrammatic representation of the word spotting framework. Fig.14. explains the combination of zone information to re-rank the word spotting results.

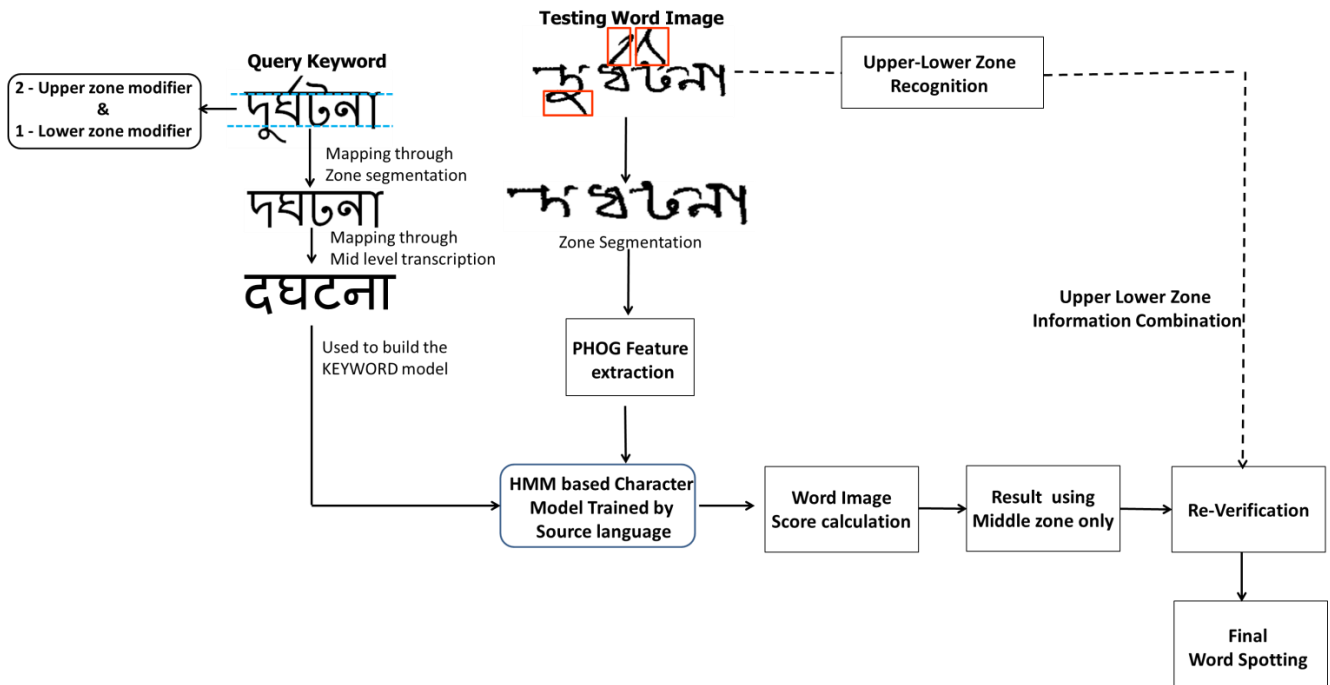

**Fig.13. Cross Language word spotting**

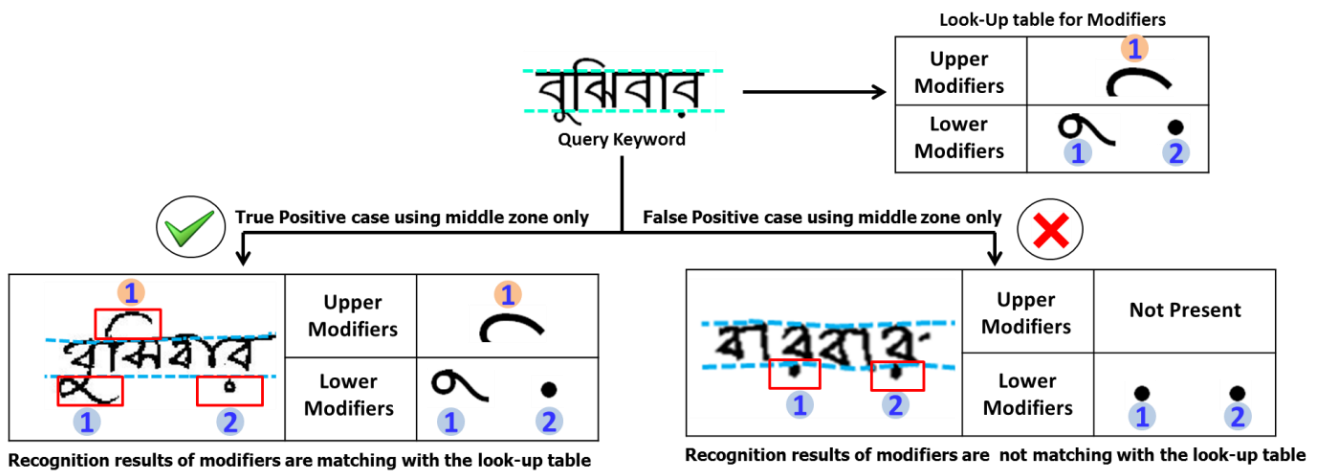

**Fig.14. True positive cases are being verified using Upper-Lower zone information and false positive cases are being eliminated using Upper-Lower zone information.**

## 5. Evaluation of Script Similarity Score

To effectively perform cross language handwriting performance evaluation, we need to determine the measure of similarity between two scripts such that character models trained using source script will perform well on target script. For this purpose, a script similarity score is needed that will measure the similarity of characters between two scripts. The score similarity score will approximately determine the extent to which target script is similar to source script.

Our proposed script similarity calculation is based on capability of recognition of characters (of target script) using source script characters. The entropy of recognition score among various characters will be used for this purpose. To find the script similarity measure between two scripts, first we create character models from source script using HMM. Next, individual characters of the target script are recognized using these source character models. During recognition process, we check for each isolated character (say, $X^T$) of the target script, the distribution of recognition probabilities among source characters. Next, we compute the entropy of $X^T$. Entropy [28], which is the measure of uncertainty, provides a higher value if the randomness to which a particular character being replaced is high and a low value if the randomness is less. The measure of entropy is given as:

$$H(X) = -\sum_{k=1}^{n} P(X_k) log_2 P(X_k) \qquad (3)$$

$$P(X_k) = \frac{Number\ of\ times\ X^T\ is\ mapped\ by\ a\ kth\ source\ character}{Total\ sample\ of\ X^T} \qquad (4)$$

Where $n$ is the number of characters of the source by which a particular test character is being replaced and $P(X_k)$ is the corresponding probability due to each replacement and given by eqn. (4). Note that, every target character will be mapped by one of the source characters.

In score similarity process, if the samples of a target script character ($X^T$) get replaced by a source script character (using majority voting as discussed in section 3.3), then it is considered that target script character is mapped better with that source character. Hence, the corresponding entropy will be less with respect to that character and this gives an indication of better cross language performance. To map the value of entropy between 0 and 1, $H_N(X)$ is normalized by following equation,

$$H_N(X) = \frac{H(X)}{1 + log_2(K)} \qquad (5)$$

where, $K$ is the total number of characters of the source script by which a particular test character is being mapped. $log_2(K)$ is the largest possible entropy [35] when all the source script characters are equiprobable for recognition of a target script character. 1 is added in denominator to avoid division by zero, which may happen when K=1. $H_N(X)$ will be maximum when the numerator H(X) is maximum. The maximum value of H(X) will be $log_2(K)$ if the samples of a character (target script) can be mapped (using recognition) by all $K$ characters of the source script. Thus, the recognition of target character is equiprobable to all source characters, which leads the maximum value of $H_N(X)$ as $\frac{log_2(K)}{1+log_2(K)}$.

The value of $H_N(X)$ is minimum when a target character is replaced always by a single source script character, i.e. $P(X_K)$ is 1 which ensures $H_N(X) = 0$. $H_N(X)$ is a measure of dissimilarity, higher of which signifies more dissimilarity and vice-versa. To convert this dissimilarity value into script similarity measurement, the similarity value of a target character ($X$) can be defined as

$$S(X) = 1 - H_N(X) \qquad (6)$$

The similarity score can be refined by including occurrences of characters in that script. It is to include weightage to characters that appear frequently. The characters of less frequency will not affect much in the cross language recognition framework compared to that of characters of high frequency. Hence, in our approach we calculated the script similarity by combining frequency of a character and the corresponding entropy of that character. The script similarity score ($S_{\text{sim}}$) is thus calculated using following equation.

$$S_{\text{sim}} = \frac{\sum_{i=0}^{M} S(X_i) \times W_i}{M} \qquad (7)$$

where $W_i$ = Frequency of occurrence of character $X_i$ and $M$ is the total number of characters in the test script. From, Eq. (2), (3) and (5), the $S_{\text{sim}}$ can be written as

$$S_{\text{sim}} = \frac{\sum_{i=0}^{M} W_i - \sum_{i=0}^{M} \frac{H(X_i)}{1 + log_2(K_i)} \times W_i}{M} \qquad (8)$$

Eq. (6) can be simplified since, $\sum_{i=0}^{M} W_i = 1$.

$$S_{\text{sim}} = \frac{1 - \sum_{i=0}^{M} \frac{H(X_i)}{1 + log_2(K_i)} \times W_i}{M} \qquad (9)$$

The script similarity $S_{\text{sim}}$ between two scripts, $S_{\text{sim(S,T)}}$ denotes the script similarity value when training has been done on source script $S$ and test characters from target script $T$. A relative script similarity index $S_{sim(S,T)}^{Rel}$ is defined by normalizing $S_{\text{sim(S,T)}}$. This is performed by dividing the score by $S_{\text{sim(T,T)}}$ when training has been done using same target script T.

$$S_{sim(S,T)}^{Rel} = \frac{S_{\text{sim(S,T)}}}{S_{\text{sim(T,T)}}} \qquad (10)$$

From experimental calculations it is observed that the score of $S_{sim(S,T)}^{Rel}$ is maximum when the source and target are similar as the amount of uncertainty between the scripts is less. The value will be less when each character of target script is replaced by characters of the source script with equal probability. The score close to 1 signifies high similarity of the two scripts while a value towards 0 signifies more uncertainty and less similarity.

---

**Algorithm 1.** Calculation of script similarity value

---

**Require:** Training data from source script and a set of isolated characters from target script

**Ensure:** Script similarity $S_{\text{sim}}$ between source and target script.

**Step 1:** All characters $(C_S{}^1 ... C_S{}^N)$ from source script are trained using HMM.

**Step 2:** Target script characters $(C_T{}^1 ... C_T{}^M)$ are recognized using training character $(C_S)$ models.

**Step 3:** Let $P_{Ti}{}^1 .. P_{Ti}{}^N$ be the recognition probabilities for target character $C_T{}^j$ with source characters. Also, let $W_i$ be the frequency of characters.

**Step 4:** Entropy is calculated using $H(X) = -\sum_{k=1}^{n} P(X_k) log_2 P(X_k)$ [from eq. (3)]

**Step 5:** Finally, script similarity value is calculated by $S_{\text{sim}} = \frac{1 - \sum_{i=0}^{M} \frac{H(X_i)}{1 + log_2(K_i)} \times W_i}{M}$ by Eq. (9)

**Step 6:** The value of $S_{\text{sim}}$ is normalized by $S_{sim(S,T)}^{Rel} = \frac{S_{\text{sim(S,T)}}}{S_{\text{sim(T,T)}}}$ using Eq. (10).

---

## 6. EXPERIMENT RESULTS AND DISCUSSION

### 6.1. Dataset Collection and Scripts used in Experiments

To the best of our knowledge, there exists no standard database to evaluate cross language handwritten text recognition and spotting tasks. To check the performance of our cross language framework, we used three Indic scripts (north Indian), namely, Devanagari, Bangla and Gurumukhi respectively. Devanagari and Bangla are two most popular Indic scripts where pieces of research work [7, 8, 13, 20, 22, 30] exist. A few datasets are available for evaluation purpose for these two scripts. In contrast, Gurumukhi is relatively a low resource script and does not have any available datasets (to our knowledge).

The dataset of Indic script [20] contains a total of 11,253(10,667) Bangla (Devanagari) word image for training and 3,856(3,589) for testing. These word images were collected from handwritten document images of individuals of different profession. A part of this dataset is collected from publicly available cmaterdb dataset [29] which contains scanned handwritten documents for both Bangla and Devanagari scripts. We also included a subset of city-name dataset [30] in Bangla script dataset. For Gurumukhi dataset, we collected a total of 40 handwritten documents written by 12 different right handed males mainly from academic profession. The words are extracted by a line segmentation method followed by word segmentation [4]. A total of 12,385 word images were extracted, out of which 9,243 word images are considered for training and rest 3,142 as testing. All the word images are manually annotated. Note that, we have not considered any conjunct characters [22] in our datasets. The dataset contains consonants, vowels and modifiers (i.e. vowels are connected to the consonant). Since, we considered zone-wise components and did not consider any consonant conjuncts, the number of unique characters is found less compared to that mentioned in [31]. The numbers of unique word in our dataset are 2152, 3981 and 2314 for Bangla, Devanagari and Gurumukhi scripts respectively. The number of word image considered for training and testing in cross-language framework is detailed in Table II. The cross language performance is tested for every combination among these three scripts. One of the scripts is used for training at a time and cross language performance is evaluated for other two scripts. We repeat this considering each script as source. The lexicons considered during the experiment are of sizes 1921, 1953, 1934 for Bangla, Gurumukhi and Devanagari respectively.

**Table II. Number of images used for training and testing**

| Source Script (as Training) | Training Word Image | Testing Word Image | | |
|---|---|---|---|---|
| | | Devanagari | Bangla | Gurumukhi |
| Devanagari | 10,667 | 3,589 | 3,856 | 3,142 |
| Bangla | 11,253 | 3,589 | 3,856 | 3,142 |
| Gurumukhi | 9,243 | 3,589 | 3,856 | 3,142 |

## 6.2. Performance of Script Similarity

In our cross language framework, the performance depends on the amount of similarity between the scripts considered as training (source) and testing (target). This similarity is evaluated based on the entropy measure between the source and target script. We have considered 3 Indic scripts namely, Bangla, Devanagari and Gurumukhi to evaluate the script similarity among each other. We have also included one Latin script, e.g. English, as one dissimilar script to check the entropy based similarity between English and three Indic scripts. Higher value of entropy ensures more dissimilarity between two scripts and results in low value of relative script similarity index $S_{sim(S,T)}^{Rel}$. Relative script similarity index is assured to be value 1 when a single script is considered for both training and testing. This extent of similarity keeps on deceasing as the relative script similarity index reduces from 1. This relative script similarity index signifies the extent of similarity between two scripts. As discussed earlier, characters in Indic scripts appear in three different zones and thus, large number of compound character units is generated through combination of vowels, modifiers and characters. Hence, we have employed the concept of zone segmentation to reduce the number of character classes and utilize the zone wise similarity among the characters. Here, we have evaluated relative script similarity index among the Indic scripts using both with and with-out zone segmentation method. Fig. 15(a) and 15(b) show the script similarity values with 4 scripts among each other. By zone segmentation method, we mean that the characters are segmented intro three zone and similarity is measure among the character units from the same zone. Characters with all three zones are used in case of without zone segmentation based method for relative script similarity evaluation. From Fig. 15(a) and 15(b), it can be inferred that degree of similarity among the different Indic scripts is much higher when we use zone segmentation method. It is due to the fact that similar structural information lies in the middle zone among the scripts. Thus, zone wise word components are used in our cross-language framework. Note that, relative script similarity index between Bangla and Devanagari is larger than between Bangla and Gurumukhi scripts. It is because characters of Bangla and Devanagari scripts share more similarity than other scripts.

| Target \ Source | Bangla | Devanagari | Gurumukhi | English |
|---|---|---|---|---|
| Bangla | 1.00 | 0.76 | 0.69 | 0.11 |
| Devanagari | 0.76 | 1.00 | 0.70 | 0.13 |
| Gurumukhi | 0.71 | 0.73 | 1.00 | 0.08 |
| English | 0.12 | 0.13 | 0.09 | 1.00 |

**(a)**

| Target \ Source | Bangla | Devanagari | Gurumukhi | English |
|---|---|---|---|---|
| Bangla | 1.00 | 0.53 | 0.46 | 0.09 |
| Devanagari | 0.54 | 1.00 | 0.49 | 0.11 |
| Gurumukhi | 0.47 | 0.49 | 1.00 | 0.07 |
| English | 0.09 | 0.12 | 0.08 | 1.00 |

**(b)**

**Fig.15. Relative script similarity index among Indic scripts (a) using zone segmentation (b) without using zone segmentation**

## 6.3. Word Recognition

We used 32 Gaussian Mixture Models and 8 states during HMM training as it provides the optimum result for our case. In our experiment, we used one of the Indic scripts as source script to train the cross language model and tested the performance for other two scripts. This process is iterated for all three scripts and results are reported. We found 74.28% and 71.54% accuracy for middle zone word recognition with top 5 choices in Bangla and Gurumukhi as target script and Devanagari as a source script. When Bangla is used as a source script, the middle zone word recognition accuracies with top 5 choices for Devanagari and Gurumukhi are found to be 75.21% and 72.03% respectively. The same for Devanagari and Bangla scripts using Gurumukhi as source script are 71.69% and 71.32% respectively. Fig. 16 shows cross language middle zone word recognition accuracies with different Top choices using Devanagari, Bangla and Gurumukhi as source script respectively.

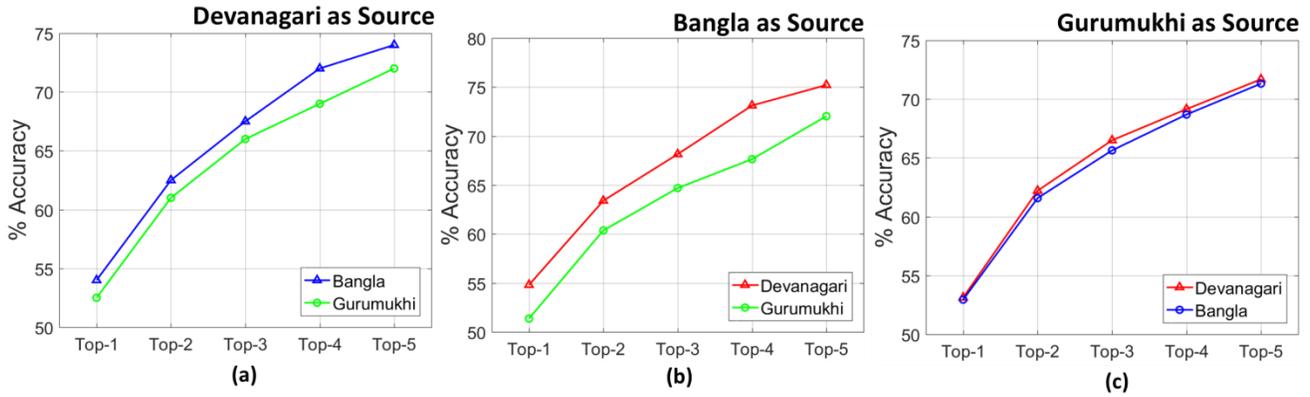

**Fig.16. Middle zone word recognition results (a) for Bangla and Gurumukhi using Devanagari as source (b) for Devanagari and Gurumukhi using Bangla as source (c) for Devanagari and Bangla using Gurumukhi as source.**

The recognition results of modifiers in upper and lower zones are given in Table III. We have collected a total of 1647, 1721 and 1494 upper zone modifiers from Devanagari, Bangla and Gurumukhi training dataset respectively. The same for lower zone modifiers are 1424, 1521, 1347 respectively. To check the performance we considered 500 modifiers of each zones for Devanagari, Bangla and Gurumukhi scripts during testing. Also, a comparative study of the performance of cross language framework with respect to traditional approach of training i.e. where training and testing is performed on the same script; in Section 5.5.

**Table III: Recognition results of the upper & lower zone modifiers by SVM.**

| | Devanagari as Source | | | |
|---|---|---|---|---|
| | **Bangla** | | **Gurumukhi** | |
| | **Top-1** | **Top-2** | **Top-1** | **Top-2** |
| **Upper zone** | 72.31 | 79.35 | 68.25 | 75.23 |
| **Lower zone** | 69.21 | 77.01 | 67.02 | 74.25 |
| | Bangla as Source | | | |
| | **Devanagari** | | **Gurumukhi** | |
| | **Top-1** | **Top-2** | **Top-1** | **Top-2** |
| **Upper zone** | 74.01 | 81.23 | 67.91 | 76.39 |
| **Lower zone** | 70.36 | 78.36 | 67.11 | 74.11 |
| | Gurumukhi as Source | | | |
| | **Devanagari** | | **Bangla** | |
| | **Top-1** | **Top-2** | **Top-1** | **Top-2** |
| **Upper zone** | 70.12 | 75.38 | 69.31 | 74.98 |
| **Lower zone** | 65.33 | 74.39 | 65.14 | 74.49 |

After getting zone-wise results from three zones, the middle zone recognition results are combined with upper and lower zone modifiers to get the final word level. Zone segmentation and combination approach gives us the flexibility of re-ranking of recognition result using the information of upper-lower zone modifiers and their corresponding position in the image. Because of this flexibility we have analysed the middle zone recognition results up to 5 top choices and considered all of them with combination of upper and lowers zone modifiers. Each possible associated word is matched with the lexicon using Levenshtein distance [34] and the lexicon word with minimum distance is considered as best result.

The combination is performed according to the alignment performed in middle zone [13, 20]. Some qualitative results are shown in Fig.17. The recognition performances at full word level are shown in Fig.18. We have achieved accuracy of 60.21% and 57.94% for top 1 in case of Bangla and Gurumukhi using Devanagari as source script respectively. The recognition performance increased to 74.28% and 71.54% considering Top 5 choices for the same. Considering Bangla as source scripts, the full word recognition result for Devanagari and Gurumukhi become 61.14% and 57.49% for top 1 respectively.

The same for Devanagari and Bangla scripts using Gurumukhi as source script are 57.77% and 57.59% respectively. We have also tested the performance using lexicon of different sizes (see Fig.19). The words in lexicon are considered arbitrarily from different newspapers.

| Word Image (Bangla) | Source Language | | Word Image (Bangla) | Source Language | |
|---|---|---|---|---|---|
| | Devanagari | Gurumukhi | | Devanagari | Gurumukhi |
| কলকাতা | ✔ কলকাতা | ✔ কলকাতা | অধিকার | ✔ অধিকার | ✔ অধিকার |
| নির্বাচন | ✔ নির্বাচন | ✘ নিবারন | কালনা | ✔ কালনা | ✔ কালনা |
| অনুরোধ | ✘ अनুরাধা | ✘ অনাবিল | মালদা | ✔ মালদা | ✘ ঝালদা |

**(a)**

| Word Image (Devanagari) | Source Language | | Word Image (Devanagari) | Source Language | |
|---|---|---|---|---|---|
| | Bengali | Gurumukhi | | Bengali | Gurumukhi |
| आपका | ✔ आपका | ✔ आपका | मजाक | ✔ मजाक | ✘ मतक |
| किसी | ✔ किसी | ✔ किसी | घुटने | ✘ घुटते | ✘ घुटनी |
| उनके | ✔ उनके | ✘ उनकी | विमान | ✔ विमान | ✔ विमान |

**(b)**

| Word Image (Gurumukhi) | Source Language | | Word Image (Gurumukhi) | Source Language | |
|---|---|---|---|---|---|
| | Devanagari | Bengali | | Devanagari | Bengali |
| উজ্জন | ✔ उज्जन | ✘ टकन | রনেগা | ✘ রনীগ | ✔ রনেগা |
| মলমল | ✔ মলমল | ✔ মলমল | দরমন | ✔ দরমন | ✔ দরমন |
| নদমীর | ✔ नদমীর | ✔ নদমীর | বিমেজ | ✘ বিমেজ | ✘ গিমেটা |

**(c)**

**Fig.17. Qualitative results for different (a) Bangla and (b) Devanagari and (c) Gurumukhi word images using cross script training. Correct and incorrect results are indicated by tick and cross labels.**

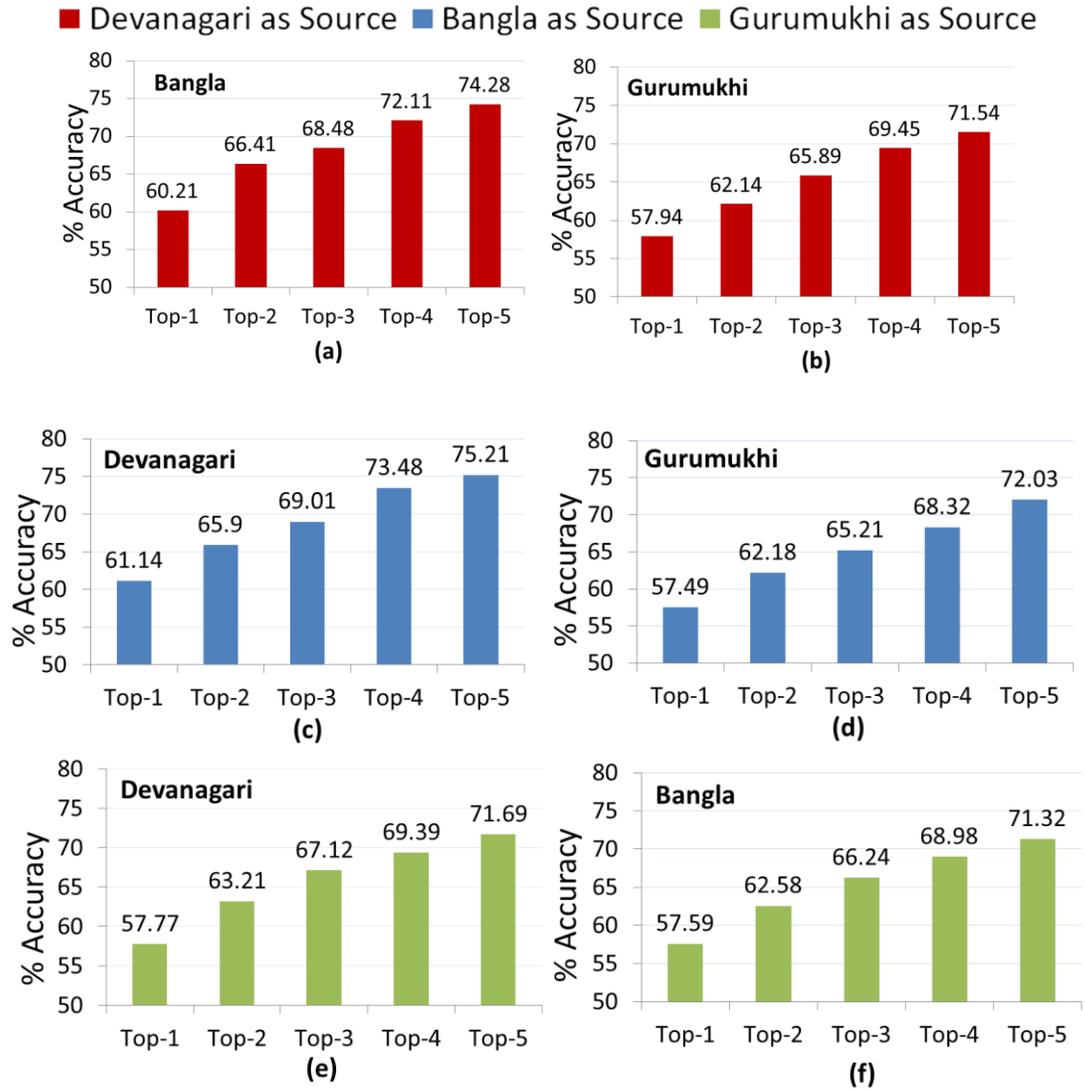

**Fig.18. Full word recognition result (considering 5 top choices) for (a) Bangla and (b) Devanagari and (c) Gurumukhi scripts using cross script training. Source script corresponding to each diagram is denoted by the color of the bar graph.**

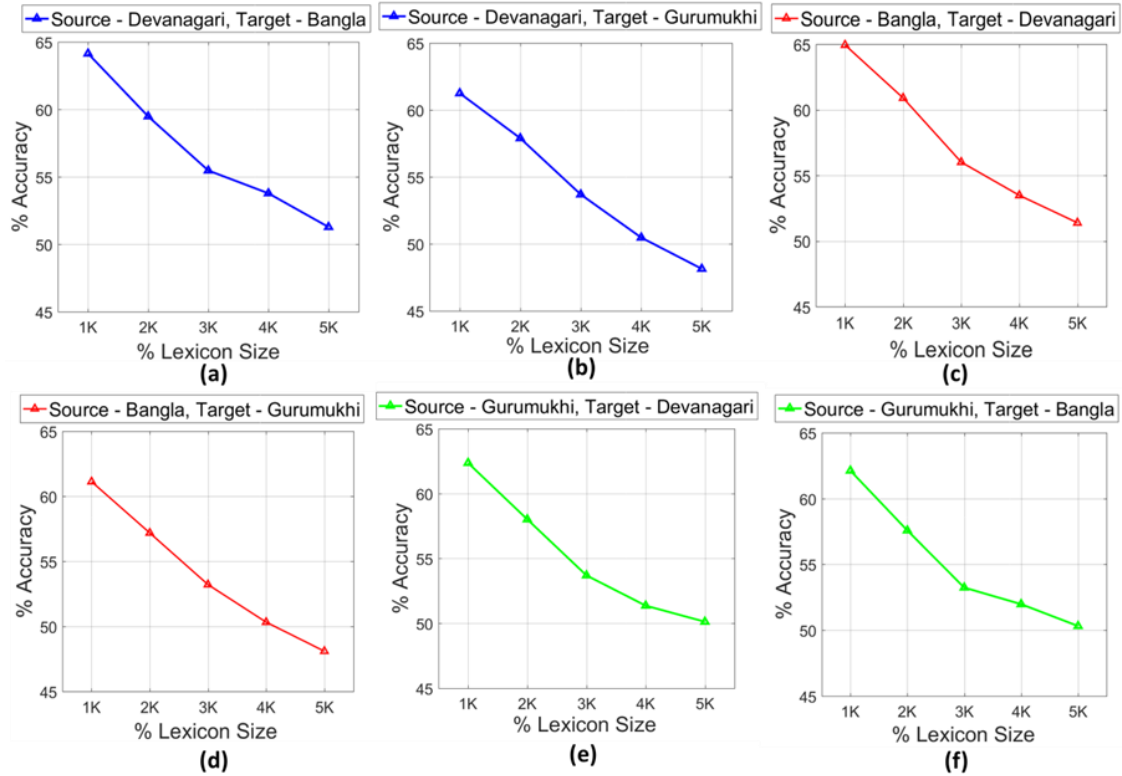

**Fig.19. Performance evaluation of word recognition using lexicons of different sizes. Source and target scripts corresponding to each diagram are mentioned in legends of the each diagram (a-f). For better viewing, we refer to the electronic version of this paper.**

### 6.4. Performance on Word Spotting

The input of our spotting system is query keyword and word images from testing script. We have measured the performance of our cross language word spotting system using precision, recall and mean average precision (MAP). The precision and recall are defined as follows.

$$Precision = \frac{TP}{TP+FN} \quad Recall = \frac{TP}{TP+FP}$$

Where, TP is true positive, FN is false negative and FP is false positive. MAP value is evaluated by the area under the curve of recall and precision.

For our experiment we noted that 32 Gaussian mixture and 8 states of HMM provided optimum results. We adopted the same method to evaluate the cross language word spotting performance as used for cross language word recognition, i.e. we used one of the Indic scripts as source scripts to train the cross language model at a time and cross language word spotting performance is evaluated for other two scripts.We have considered a total of 200 query words for each of the scripts to evaluate the performance of cross language word spotting method. Qualitative results are shown in Fig. 20 for Bangla, Gurumukhi and Devanagari word images using cross language training. Word spotting performance for middle zoned images and combination of upper-lower zone information is shown by the precision recall curve in Fig. 21 using global threshold [19]. We have obtained global mean average precision of 68.01 (67.21) and global average recall of 67.14 (66.08) for Bangla (Gurumukhi) as the target (or testing) script and Devanagari as the source script. When Bangla is used as the source script, the global average mean precision and global average recall were found to be 68.94 (66.79) and 68.04 (66.19) respectively for Devanagari (Gurumukhi) script. The same for Devanagari (Bangla) script using Gurumukhi as the source script were found to be 66.87(66.42) and 66.14(65.97) respectively.

| Query keyword (Bangla) | Word Image | Devanagari as Source | | Gurumukhi as Source | | Query keyword (Bangla) | Word Image (Bangla) | Devanagari as Source | | Gurumukhi as Source | |
|---|---|---|---|---|---|---|---|---|---|---|---|
| | | (i) | (ii) | (i) | (ii) | | | (i) | (ii) | (i) | (ii) |
| অনুরাধা | *(word image)* | ✗ | ✓ | ✓ | ✓ | অভিযোগ | *(word image)* | ✗ | ✓ | ✗ | ✓ |
| সংবাদদাতা | *(word image)* | ✓ | ✓ | ✗ | ✗ | অনুরোধ | *(word image)* | ✗ | ✓ | ✗ | ✗ |
| বিধানভবন | *(word image)* | ✗ | ✗ | ✗ | ✗ | রাজনীতি | *(word image)* | ✗ | ✓ | ✗ | ✓ |

(a)

| Query keyword (Gurumukhi) | Word Image | Devanagari as Source | | Bangla as Source | | Query keyword (Gurumukhi) | Word Image (Gurumukhi) | Devanagari as Source | | Bangla as Source | |
|---|---|---|---|---|---|---|---|---|---|---|---|
| | | (i) | (ii) | (i) | (ii) | | | (i) | (ii) | (i) | (ii) |
| ਪੰਜਾਬੀ | *(word image)* | ✗ | ✓ | ✗ | ✓ | ਸਾਜਨ | *(word image)* | ✗ | ✓ | ✗ | ✓ |
| ਭਜਨ | *(word image)* | ✓ | ✓ | ✓ | ✓ | ਸਮੀਖਿਆ | *(word image)* | ✗ | ✗ | ✗ | ✓ |
| ਵਿਗਿਆਨ | *(word image)* | ✗ | ✓ | ✗ | ✗ | ਹਿਸਾਬ | *(word image)* | ✗ | ✓ | ✓ | ✓ |

(b)

| Query keyword (Devanagari) | Word Image | Bangla as Source | | Gurumukhi as Source | | Query keyword (Devanagari) | Word Image | Bangla as Source | | Gurumukhi as Source | |
|---|---|---|---|---|---|---|---|---|---|---|---|
| | | (i) | (ii) | (i) | (ii) | | | (i) | (ii) | (i) | (ii) |
| जिसमें | *(word image)* | ✗ | ✓ | ✗ | ✓ | मौजूदा | *(word image)* | ✗ | ✓ | ✗ | ✓ |
| तमिलनाडु | *(word image)* | ✗ | ✓ | ✗ | ✗ | विमान | *(word image)* | ✓ | ✓ | ✓ | ✓ |
| समेटने | *(word image)* | ✓ | ✓ | ✗ | ✗ | यूवाओं | *(word image)* | ✗ | ✓ | ✗ | ✓ |

(c)

**Fig.20: Example showing qualitative word spotting performance (i) without using zone segmentation (ii) using zone segmentation for (a) Bangla and (b) Gurumukhi (c) Devanagari scripts where cross language training method is utilized.**

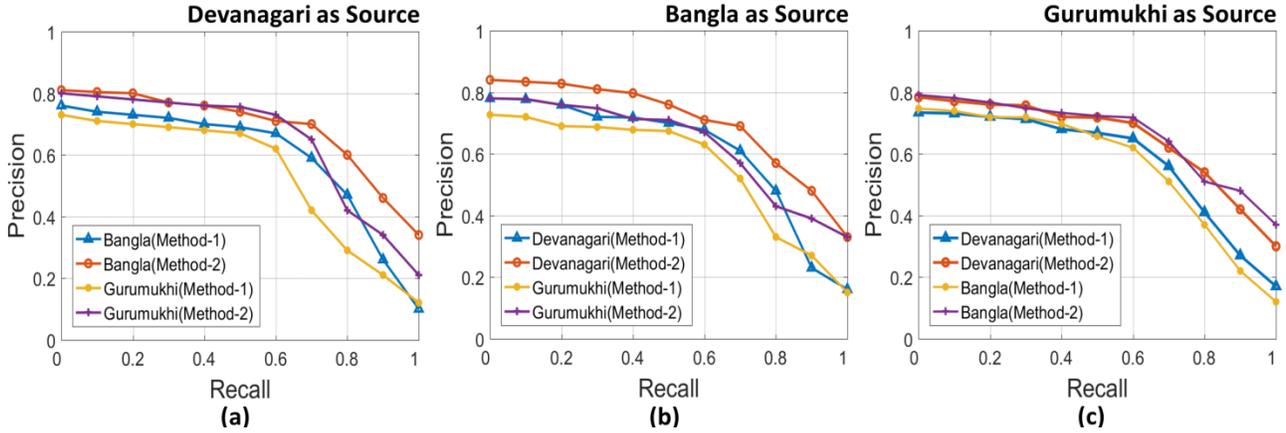

**Fig.21: Comparison of word spotting performance using middle zone only (denoted as method 1) and modifiers combination (denoted as method 2) using (a) Devanagari (b) Bangla and (c) Gurumukhi as source script respectively. Target scripts are mentioned at the legends of each diagram. For better viewing, we refer to the electronic version of this paper.**

We evaluated precision-recall curve using different number of query words (keywords) (See Fig. 22). The global MAP values are evaluated for different length of keywords and a curve has been plot in Fig. 23 to show the performance for keywords of variable length. The successive improvement in the MAP values is obtained due to implementation of zone segmentation based approach over full zoned based recognition. The improvement due to information combination from upper-lower zone modifiers is given in Table IV. From Table IV, it can be inferred that a significant improvement has been found due to inclusion of zone segmentation method in our cross language word spotting framework. Here, local MAP value signifies that single image has been considered for optimization of the threshold value whereas a global value has been used for all query keyword in case of standard MAP value evaluation.

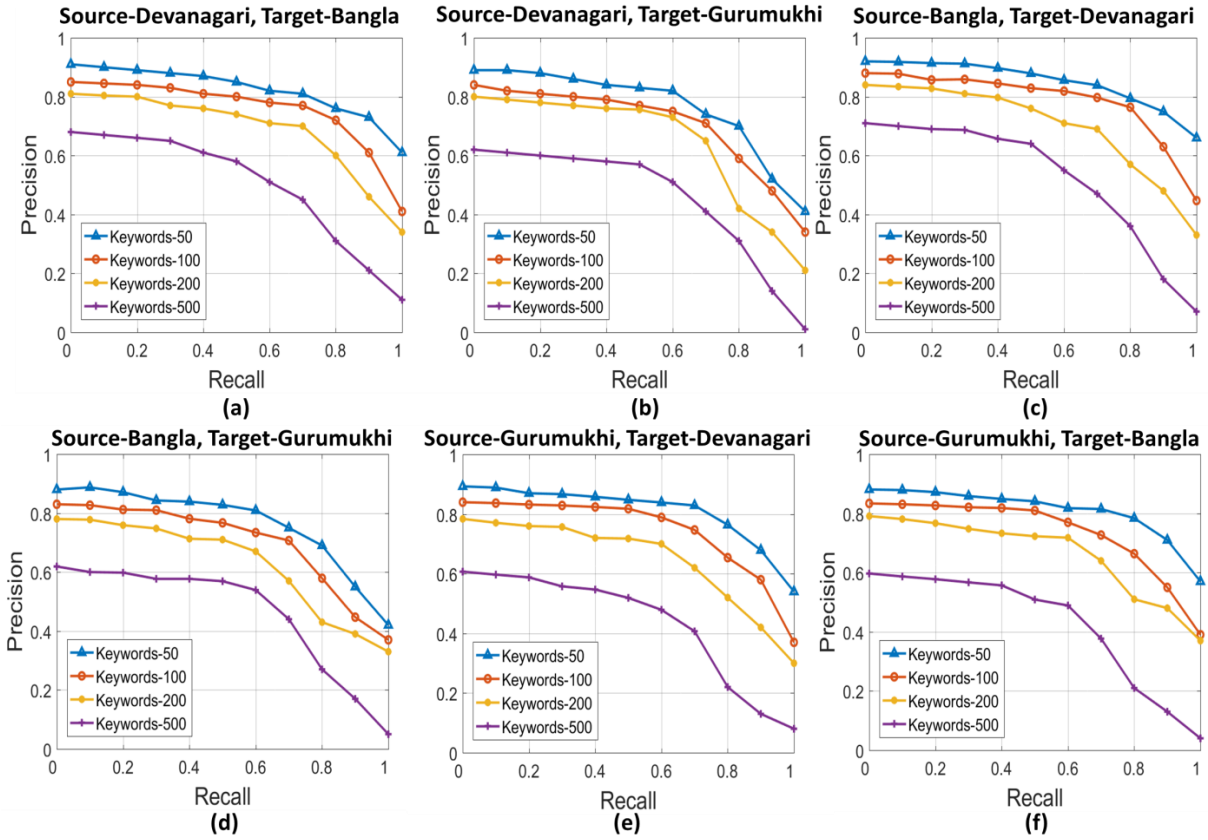

**Fig. 22: Comparative study of word spotting performance with different number of keywords. Source and target scripts are mentioned at the top of each diagram correspondingly.**

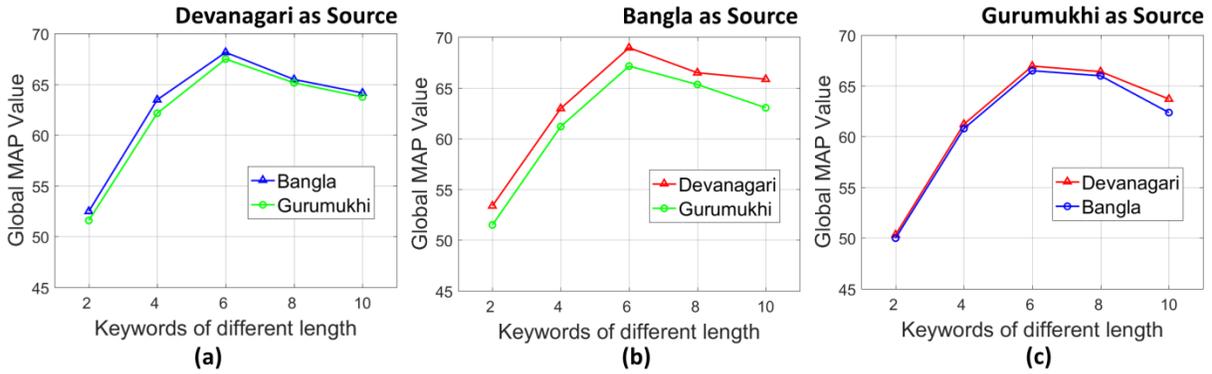

**Fig. 23: Word spotting performance using keywords of different length using (a) Devanagari (b) Bangla and (c) Gurumukhi as source script respectively. Target scripts are mentioned at the legends of each diagram. For better viewing, we refer to the electronic version of this paper.**

**Table IV. MAP values using different method**

| Approach | Threshold | Devanagari as Source | | Bangla as Source | | Gurumukhi as Source | |
|---|---|---|---|---|---|---|---|
| | | Bangla | Gurumukhi | Devanagari | Gurumukhi | Devanagari | Bangla |
| Without using zone segmentation | Local | 49.56 | 50.17 | 50.21 | 50.32 | 49.02 | 48.67 |
| | Global | 41.14 | 41.68 | 42.36 | 41.84 | 40.91 | 40.21 |
| Using Zone segmentation (middle zone only) | Local | 68.14 | 67.58 | 68.83 | 67.94 | 66.18 | 65.98 |
| | Global | 60.14 | 59.44 | 61.23 | 60.12 | 58.14 | 57.69 |
| Combination of middle zone and upper-lower zone modifiers | Local | 76.14 | 76.12 | 77.08 | 76.52 | 74.68 | 74.31 |
| | Global | **67.14** | **66.57** | **67.48** | **66.51** | **66.51** | **66.21** |

## 6.5. Parameter Evaluation

A comprehensive study is performed to find the optimum value of parameters used in our cross language framework. We used continuous density HMMs with diagonal covariance matrices of GMMs in each state. We evaluated both our cross language word recognition and word spotting framework with varying Gaussian mixtures (16, 32, 64, 128 and 256) and state numbers (6, 7, 8, and 9). For our experiment, we found that 32 Gaussian mixture and 8 states of HMM Training provided optimum results for both cross language word spotting and recognition. Table V shows the cross-language performance using varying Gaussian mixtures and state numbers. Also, the upper and lower zone modifiers are tested using SVM at different values of cost parameter (C). These values include the range from 0.1 to 1 with an interval of 0.1, from 1 to 10 with an interval of 1 which is followed by the range 10 to 100 with an interval of 10. The optimum value has been found experimentally as 1.

**Table V. Cross language word recognition (% accuracy) and word spotting (global MAP value) performance using varying Gaussian number and State number**

| | Gaussian Number | Devanagari as Source | | Bangla as Source | | Gurumukhi as Source | |
|---|---|---|---|---|---|---|---|
| | | Bangla | Gurumukhi | Devanagari | Gurumukhi | Devanagari | Bangla |
| **Word Recognition** | 16 | 58.14 | 55.91 | 58.79 | 55.47 | 55.17 | 55.14 |
| | 32 | **60.21** | **57.94** | **61.14** | **57.49** | **57.77** | **57.59** |
| | 64 | 59.12 | 56.10 | 59.64 | 56.12 | 56.19 | 56.01 |
| | 128 | 53.17 | 51.69 | 54.97 | 51.69 | 52.36 | 51.96 |
| | 256 | 48.96 | 47.64 | 49.96 | 47.69 | 47.69 | 47.57 |
| | **State Number** | **Devanagari as Source** | | **Bangla as Source** | | **Gurumukhi as Source** | |
| | | Bangla | Gurumukhi | Devanagari | Gurumukhi | Devanagari | Bangla |
| | 6 | 58.48 | 55.91 | 59.47 | 55.17 | 55.91 | 55.79 |
| | 7 | 59.14 | 56.19 | 60.91 | 56.41 | 56.47 | 56.67 |
| | 8 | **60.21** | **57.94** | **61.14** | **57.49** | **57.77** | **57.59** |
| | 9 | 59.49 | 56.14 | 60.19 | 56.19 | 56.61 | 56.71 |
| **Word Spotting** | **Gaussian Number** | **Devanagari as Source** | | **Bangla as Source** | | **Gurumukhi as Source** | |
| | | Bangla | Gurumukhi | Devanagari | Gurumukhi | Devanagari | Bangla |
| | 16 | 65.39 | 64.94 | 64.94 | 63.84 | 64.84 | 64.78 |
| | 32 | **67.14** | **66.57** | **67.48** | **66.51** | **66.51** | **66.21** |
| | 64 | 66.14 | 65.12 | 66.39 | 65.48 | 65.47 | 65.84 |
| | 128 | 63.17 | 61.47 | 63.64 | 62.19 | 61.49 | 61.57 |
| | 256 | 60.96 | 59.48 | 61.47 | 58.48 | 59.79 | 59.61 |
| | **State Number** | **Devanagari as Source** | | **Bangla as Source** | | **Gurumukhi as Source** | |
| | | Bangla | Gurumukhi | Devanagari | Gurumukhi | Devanagari | Bangla |
| | 6 | 65.49 | 64.14 | 65.94 | 64.97 | 64.87 | 64.76 |
| | 7 | 66.94 | 65.91 | 66.74 | 65.94 | 65.48 | 65.54 |
| | 8 | **67.14** | **66.57** | **67.48** | **66.51** | **66.51** | **66.21** |
| | 9 | 66.59 | 66.13 | 66.69 | 65.19 | 65.39 | 65.61 |

## 6.6. Comparison with traditional training approach

To the best of our knowledge, there exist no earlier works dealing with cross-lingual word recognition and spotting. To measure the performance of our cross language framework, we compare it with traditional word recognition/spotting method where training and testing are done on the same script. The number of word image considered for training and testing in each set of combination is mentioned in the Table II. The lexicons considered are of sizes 1921, 1953 and 1934 for Bangla, Gurumukhi and Devanagari respectively. For word spotting, we considered a total of 200 query keyword for each experiment. Results of word recognition and word spotting are given in Fig. 24 and Fig. 25. Here, we summarize the results of possible combinations of cross-language framework among these three scripts. Fig 26 Fig. 27 show comparative study of cross language framework with traditional training approach.

| Target / Source | Bangla | Devanagari | Gurumukhi |
|---|---|---|---|
| Bangla | 82.18 | 61.14 | 57.49 |
| Devanagari | 60.21 | 83.97 | 57.94 |
| Gurumukhi | 57.59 | 57.77 | 82.01 |

**Fig.24. Word recognition accuracy using different script combination**

| Target / Source | Bangla | Devanagari | Gurumukhi |
|---|---|---|---|
| Bangla | 72.45 | 67.48 | 66.01 |
| Devanagari | 67.14 | 73.18 | 66.57 |
| Gurumukhi | 66.21 | 66.51 | 72.11 |

**Fig.25. Global MAP values of word spotting for using different script combination**

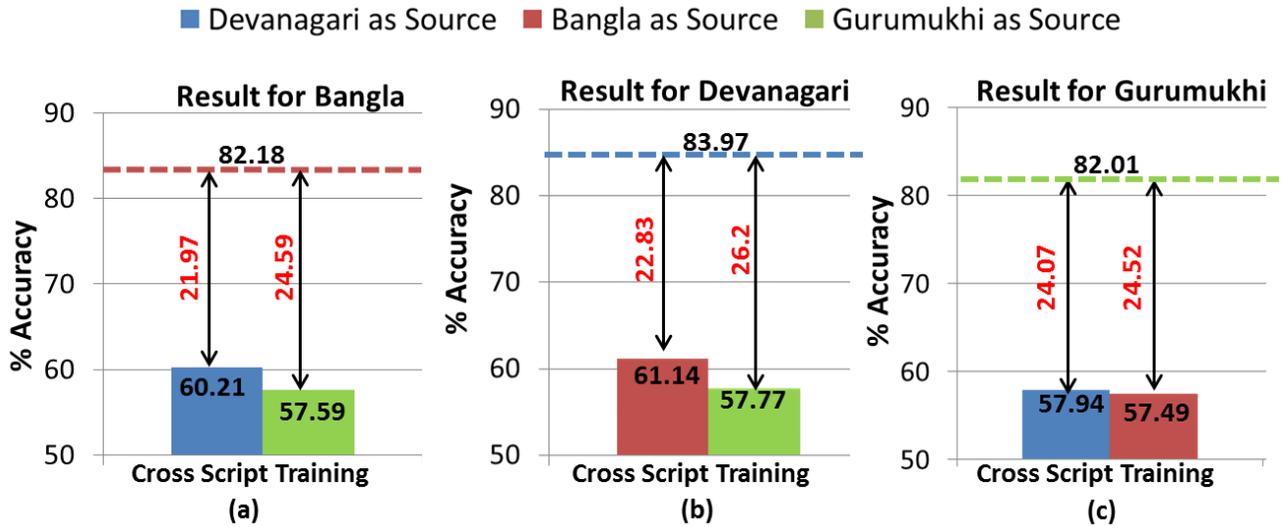

**Fig.26. Comparative study of cross language word recognition framework where dotted horizontal line in each diagram shows the recognition accuracy obtained through traditional training approach. The recognition performance corresponding to different source scripts are shown using color bars. For better viewing, we refer to the electronic version of this paper.**

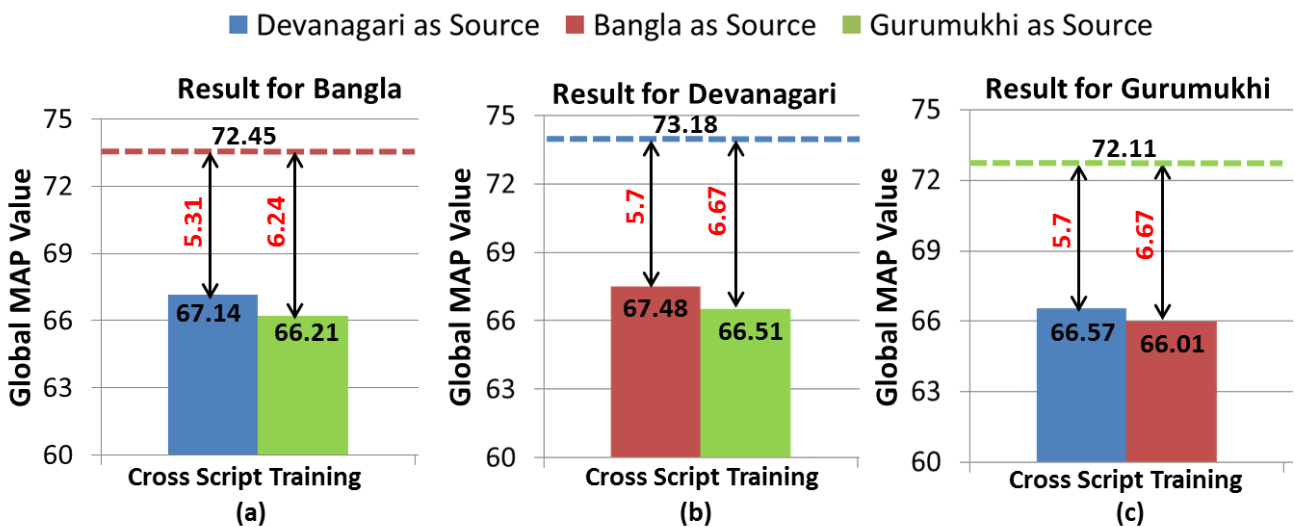

**Fig.27. Comparative study of cross language word spotting framework where dotted horizontal line in each diagram shows the Global MAP value obtained through traditional training approach. The MAP values corresponding to different source scripts are shown using color bars. For better viewing, we refer to the electronic version of this paper.**

**6.7: Error analysis** While mapping the characters from source to target script, although preference was given according to majority voting, it was noted that some characters from source script are matched very closely to target script during majority voting, hence confusion occurred. Thus, some words are being recognized wrongly. Also, we noticed that during lexicon matching process more than one substitute can be possible during character mapping. Say for example Bangla character 'উ' and 'ত' are both mapped with Devanagari character 'उ'. So sometimes a word is recognized wrongly in those particular cases. Such errors are shown in Fig.28. We noted that such confusion also affect the word spotting system. Fig. 29 shows qualitative results of cross language word recognition and word spotting on full Bangla text line images. Fig. 29(a) and Fig. 29(b) show the recognition results of a Bangla text line using training with 3 Indic scripts, Bangla, Gurumukhi and Devanagari. Note that, though few words were not recognized properly the overall recognition performance is encouraging. Similarly, we show word spotting results using cross-language framework in Fig. 29(c). Two query words were searched in the dataset with training with different source scripts and the results were marked in bounding box in those text lines.

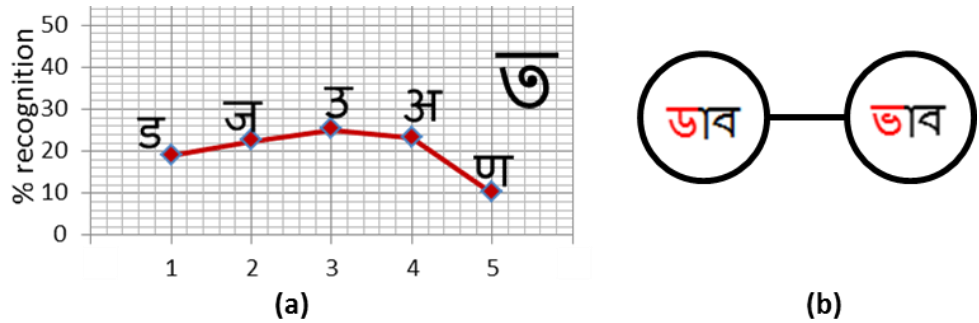

**(a)**          **(b)**

**Fig.28. (a) Close recognition rate of a character during majority voting. (b) More than one substitutes for lexicon matching.**

| Source/Training Language | _অকদিন ঋতুপর্ণ ঘোস হঠাং আমাকে ফোন করে বললেন_ |
|---|---|
| Bangla | একদিন ঋতুপর্ণ ঘা**স** হঠাৎ আমাকে ফোন করে বললেন |
| Gurumukhi | এক**লা** ঋতুপর্ণ **ঘাস** হঠা**ন** আমাকে **লোন** করে বললেন |
| Devanagari | এক**দা** ঋতুপর্ণ ঘোষ হঠাৎ **তা**মাকে **কোন** করে বললেন |

**(a)**

| Source/Training Language | _মানুসের জীবনটা পৃথিবীর নানা জীবের ইতিহাসের_ |
|---|---|
| Bangla | মানুষের জীবনটা পৃথিবীর নানা জীবের ইতিহাসের |
| Gurumukhi | **ফা**নু**সে**র জীবনটা পৃথিবীর **মা**না **ডা**বের ইতিহাসের |
| Devanagari | মানু**সে**র জীবনটা পৃথিবীর না**চা** জীবের ইতিহাসের |

**(b)**

| Query Keyword | Text Line Image | Source/Training Language | | |
|---|---|---|---|---|
| | | Bangla | Devanagari | Gurumukhi |
| ফুটপাত | (text line image) | ✓ | ✓ | ✗ |
| | (text line image) | ✓ | ✓ | ✓ |
| | (text line image) | ✓ | ✗ | ✗ |
| কলম | (text line image) | ✓ | ✓ | ✓ |
| | (text line image) | ✓ | ✗ | ✗ |
| | (text line image) | ✓ | ✗ | ✓ |

**(c)**

**Fig.29: (a) and (b) show full sentence recognition result for Bangla text lines using different scripts for training (wrong recognition is indicated by red mark). (c) Word spotting performance for Bangla text lines using different scripts for training and result is given by correct (by tick) and incorrect (by cross) label.**

## 7. Conclusion & Future work

In this paper we propose a novel method for cross-language handwritten text recognition and word spotting. There are many languages in India for which handwritten text recognition systems have not been explored due to lack of proper training data. Our approach deals with handwritten recognition of these low resource scripts where we use a script with higher number of available samples to train and test the script with lower available samples. The criteria for selecting the script with higher number of samples to train the system for a particular low resource script depends on the script similarity score between the two scripts. The script similarity score indicates the accuracy to be obtained on testing with the low resource script. A higher script similarity score will give a better performance. Thus based on this script similarity score a character mapping was performed. Word spotting was also performed using this cross language approach. This is the first work of its kind and we hope this cross-language work will be a step forward for recognizing other such low-resource scripts.

In this present work, we did not consider any consonant conjuncts in datasets. Thus, the script similarity scores of every script-pair may get reduced if consonant conjuncts are taken in account. Also, one of the limitations of the proposed framework is that the performance depends on the quality of zone segmentation output [20]. If the words are not properly segmented using zone segmentation method, the cross language framework may not work. These issues could be considered in future studies. However, a combination of with and without zone segmentation based method may be considered to avoid such limitation. In future we will work on a better character mapping approach to increase the recognition efficiency. This may improve the cross language recognition performance when the lexicon size is more. Also, we will test our framework in different non-Indic handwritten scripts.